\documentclass[conference]{IEEEtran}
\IEEEoverridecommandlockouts
\usepackage{cite}
\usepackage{amsmath,amssymb,amsfonts}
\usepackage{graphicx}
\usepackage{caption}
\usepackage{subcaption}
\usepackage{textcomp}
\usepackage{xcolor}
\usepackage{algorithm}
\usepackage{algpseudocode}
\def\BibTeX{{\rm B\kern-.05em{\sc i\kern-.025em b}\kern-.08em
    T\kern-.1667em\lower.7ex\hbox{E}\kern-.125emX}}
\begin{document}

\title{FedSSC: Shared Supervised-Contrastive Federated Learning\\
}

\author{
\IEEEauthorblockN{Sirui Hu*}
\IEEEauthorblockA{\textit{SEAS} \\
\textit{Harvard University}\\
siruihu\\
@g.harvard.edu}
\and
\IEEEauthorblockN{Ling Feng*}
\IEEEauthorblockA{\textit{HSPH} \\
\textit{Harvard University}\\
lingfeng\\
@hsph.harvard.edu}
\and
\IEEEauthorblockN{Xiaohan Yang*}
\IEEEauthorblockA{\textit{SEAS} \\
\textit{Harvard University}\\
xiaohan\_yang\\@g.harvard.edu}
\and
\IEEEauthorblockN{Yongchao Chen*}
\IEEEauthorblockA{\textit{SEAS} \\
\textit{Harvard University}\\
yongchaochen\\@fas.harvard.edu}
}

\maketitle

\begin{abstract}
Federated learning is widely used to perform decentralized training of a global model on multiple devices while preserving the data privacy of each device. However, it suffers from heterogeneous local data on each training device which increases the difficulty to reach the same level of accuracy as the centralized training. Supervised Contrastive Learning which outperform cross-entropy tries to minimizes the difference between feature space of points belongs to the same class and pushes away points from different classes. We propose Supervised Contrastive Federated Learning in which devices can share the learned class-wise feature spaces with each other and add the supervised-contrastive learning loss as a regularization term to foster the feature space learning. The loss tries to minimize the cosine similarity distance between the feature map and the averaged feature map from another device in the same class and maximizes the distance between the feature map and that in a different class. This new regularization term when added on top of the moon regularization term is found to outperform the other state-of-the-art regularization terms in solving the heterogeneous data distribution problem.  
\end{abstract}


\section{Introduction}
Federate Learning \cite{zhang2022fine, zhu2021data, li2020federated, chen2022anomalous, huang2022learn} has become a hot research topic in recent years due to applications in many fields where participants don't want to share private training data but want to have a co-trained model with high quality. However, the usually non-iid distribution of training data of participating clients harms the model convergence and model accuracy to a large extent. Supervised Contrastive Learning which tries to minimize the difference between feature space of points belonging to the same class and pushes away points from different classes was found to outperform cross-entropy \cite{simclr,simsiamese}. MOON \cite{moon} is one of the state-of-the-art regularization terms that is found to be very effective in solving heterogeneity problem in federated learning by utilizing the similarity between model representations to correct the local training of individual parties. Inspired by these two ideas, we propose Shared Supervised-Contrastive Federated Learning(FedSSC) to tackle the heterogeneity problem. In FedSSC, devices can share with each other the learned class-wise feature embeddings and add the supervised-contrastive learning loss as a regularization term to foster the feature space learning. The loss tries to minimize the cosine similarity distance between the current sample's feature embedding and the averaged feature embedding from another device if they are in the same class, and on the contrary maximizes the distance if they are in different classes. This new regularization term when added on top of the MOON regularization term is found to outperform the other state-of-the-art regularization terms by getting higher accuracy and converging in fewer rounds.  

\section{Problem to Solve}
\subsection{Problem Statement}
We are particularly interested in the non-IID setting of federated learning. Specifically, we assume that there are $N$ devices, and each of them has the training data $D_i$ where $i \in \{1, 2, ..., N\}$. Each $D_i$ consists of a different distribution of class labels. Our goal is to learn a machine learning model without local devices directly sharing training data to solve

\begin{equation} \label{eq:loss term}
argmin_{w} L(w) = \sum_i^N \frac{|D_i|}{|D|} L_i(w)
\end{equation}
 where $L_i(w)$ is the empirical loss of the local device.

\subsection{Background}
Federated learning is widely used to perform decentralized training of a global model on multiple devices while preserving the data privacy of each device. One of the most basic yet popular models is FedAvg \cite{fedavg} algorithm, in which a central server aggregates the local model weights on each device to build a global model without directly accessing the local training data. However, it suffers from heterogeneous local data on each training device which increases the difficulty of reaching the same level of accuracy as the centralized training. To tackle the heterogeneity, MOON proposes to add a regularization term to 
prevent local feature representation of the image from being too far from the global feature representation of the same image \cite{moon}. MOON outperformed other regularization terms like FedProx \cite{fedprox}.

\subsection{Research Goal}
The goal of our research is to improve the federated learning algorithm in non-IID scenarios without sharing raw data across devices. Specifically, we would like to introduce an extra regularization for class-wise feature space through supervised contrastive loss on top of the MOON regularization term.

\subsection{Evaluation Metrics}
To compare our method with other existing approaches, we evaluate the model performance by the top-1 accuracy of global model on an isolated test set. Moreover, we use the number of communication rounds to achieve the same level of accuracy as our metric for  convergence speed.

\section{Approach}
Inspired by the idea of Supervised Contrastive Learning, in addition to the MOON loss FedSSC utilizes class-wise average feature maps shared by other devices to correct local training and to tackle the heterogeneity problem. The objective function for the local device is composed of three parts: 1) typical supervised learning loss term calculated with cross-entropy ($l_{class}$), 2) MOON loss ($l_{moon}$), and 3) global class-wise contrastive loss ($l_{glob}$). With $\tau$ being the temperature, $l_{moon}$ takes the projected feature representations $z$, $z_{glob}$, and $z_{prev}$ by passing the same image into the current model, the current round's global model, and the previous epoch's model\cite{moon}.  
\begin{equation} \label{eq:moon_loss}
l_{moon}= -log \frac{exp(sim(z,z_{glob})/\tau)}{exp(sim(z,z_{glob})/\tau)+exp(sim(z,z_{prev})/\tau)}\\
\end{equation}

Similarly $l_{glob}$ takes the temperature $\tau$, the current projected feature representations $z^i$ in class $i$ and the shared global class-wise projected feature representations $zs_{glob}$ with a total of $|K|$ classes. For the shared global representation in the same class as the image, we treat them as a positive pair, whereas any other shared global representations in a different class as negative pairs. 
\begin{equation} \label{eq:global_loss}
l_{glob}= -log \frac{exp(sim(z^i,zs^i_{glob})/\tau)}{\sum_{k \in K} exp(sim(z^i,zs^k_{glob})/\tau)}\\
\end{equation}

To construct the $zs_{glob}$, we first have each local device report to global server the class-wise projected feature representations at the last round of epoch using Equation \ref{eq:global_rep} assuming that each device has $N$ training data with $N^k$ of them in class $k$. In each communication round, for each class the global server randomly selects a device who has at least 10 images of that class locally as a source of that class's class-wise feature representation.

\begin{equation} \label{eq:global_rep}
zs^k_{glob} = \frac{\sum_{j}^N z^i_j * 1_{i=k}}{N^k}
\end{equation}

As shown in Equation \ref{eq:loss}, we can tune the two parameters $\mu_{moon}$ and $\mu_{glob}$ to weight the MOON loss and global class-wise contrastive loss differently. The local objective is to minimize the $l$.
\begin{equation} \label{eq:loss}
l = l_{class} + \mu_{moon}*l_{moon} + \mu_{glob} * l_{glob}\\
\end{equation}

\begin{algorithm}
\caption{FedSSC Framework}\label{alg:cap}
\begin{algorithmic}
 \renewcommand{\algorithmicrequire}{\textbf{Input:}}
  \renewcommand{\algorithmicensure}{\textbf{Output:}}
\Require number of communication rounds $T$,
number of devices $P$, number of local
epochs $E$, temperature $\tau$ , learning rate $\eta$,
hyper-parameter $\mu$, total number of data $N$, number of data in device$_i$ is  $N_i$
\Ensure Final global model $w^T$
\\ \textit{\textbf{Global Server}} :
\\ \text{Initialize global model $w^0$}
\\ \text{and classwise feature representation $zs^0$}
\For{$t=0,1,...,T-1$}
    \For{$i=1,...,P$}
        \State send the $w^t$ and the $zs^t$ to device $i$
        \State $w_i^{t+1} , zs_i^{t+1} \gets$ \textbf{LocalTraining}($i,w^t,zs^{t}$)
    \EndFor
    \State $w^{t+1}\gets \sum_i^P \frac{|N_i|}{|N|} w_i^t$
    \State $zs^{t+1}\gets \sum_i^P \frac{zs_i^{t+1}}{P} $
\EndFor
\State return $w^T$
\\
\\ \textit{\textbf{LocalTraining}} :
\State $w_0^t = w^t$
\For{$i=0,...,E-1$}
    \For{each batch b = ($x,y$) of $N_i$}
    \State $l_{class} \gets CrossEntropyLoss(F_{w_i^t}(x),y)$
    \State $z \gets Proj(Enc(w_i^t;x))$
    \State $z_{glob} \gets Proj(Enc(w^t;x))$
    \State $z_{prev} \gets  Proj(Enc(w_{i}^{t-1};x))$
    \State $l \gets  l_{class} + \mu_{moon}*l_{moon}(z,z_{glob},z_{prev}) + \mu_{glob} * l_{glob}(z,zs^t)$
    \State $w_{i+1}^t \gets w_i^t - \eta \Delta l $
    \EndFor
\EndFor
\For{each class c $\in$ C}
    \State $zs^c = \frac{1}{N^c}\sum_{x}^N Proj(Enc(w_E^t;x^j)) * 1_{j=c}$
\EndFor
\State return $w_E^{t+1}$, $zs$\\
\end{algorithmic}
\end{algorithm}

\section{Intellectual Points}
Our contributions are two-fold. First, most prior works focused on regularizing local devices' weights or only regularizing feature representations of local images\cite{moon},\cite{fedprox}. However, our approach directly takes advantage of feature representations from other devices without sharing the raw data by regularizing each sample's local representation with its corresponding global class-wise feature representation. The global class-wise feature map is from a randomly selected device for each round

Furthermore, our experiments show that the global representation contrastive loss and the MOON loss are complementary to each other. We have considered losses other than the MOON loss by modifying its negative pair, but the model could easily collapse or the performance would be worse than the FedAvg. Moreover, even without the MOON loss, simply adding our regularization term on top of the supervised learning loss can achieve the same level of accuracy with MOON. However, using one of them doesn't outperform the other. Combining them together is the key to our success.  

\section{Work Performed}
\subsection{Dataset}
We used CIFAR-10 as our experiment dataset because it is relatively small and widely used in previous papers. To simulate the non-IID scenario, we follow MOON's setup by using the Dirichlet distribution to generate dataset $D_i$ for each device. Specifically, we sample ${p_k} \sim Dir_N(\beta)$ for each class and allocate $p_{kj}$ samples of class $k$ to device $j$. Our default is $\beta=0.5$, which simulates a severe non-IID situation. The larger the $\beta$ gets to, the more IID each device will be. Using Dirichlet, we can have each local device have the same total number of samples as each other, but different class distribution from each other.

\begin{figure*}[ht]
\centering
\begin{subfigure}{0.24\linewidth}
  \centering
  \includegraphics[width=\linewidth]{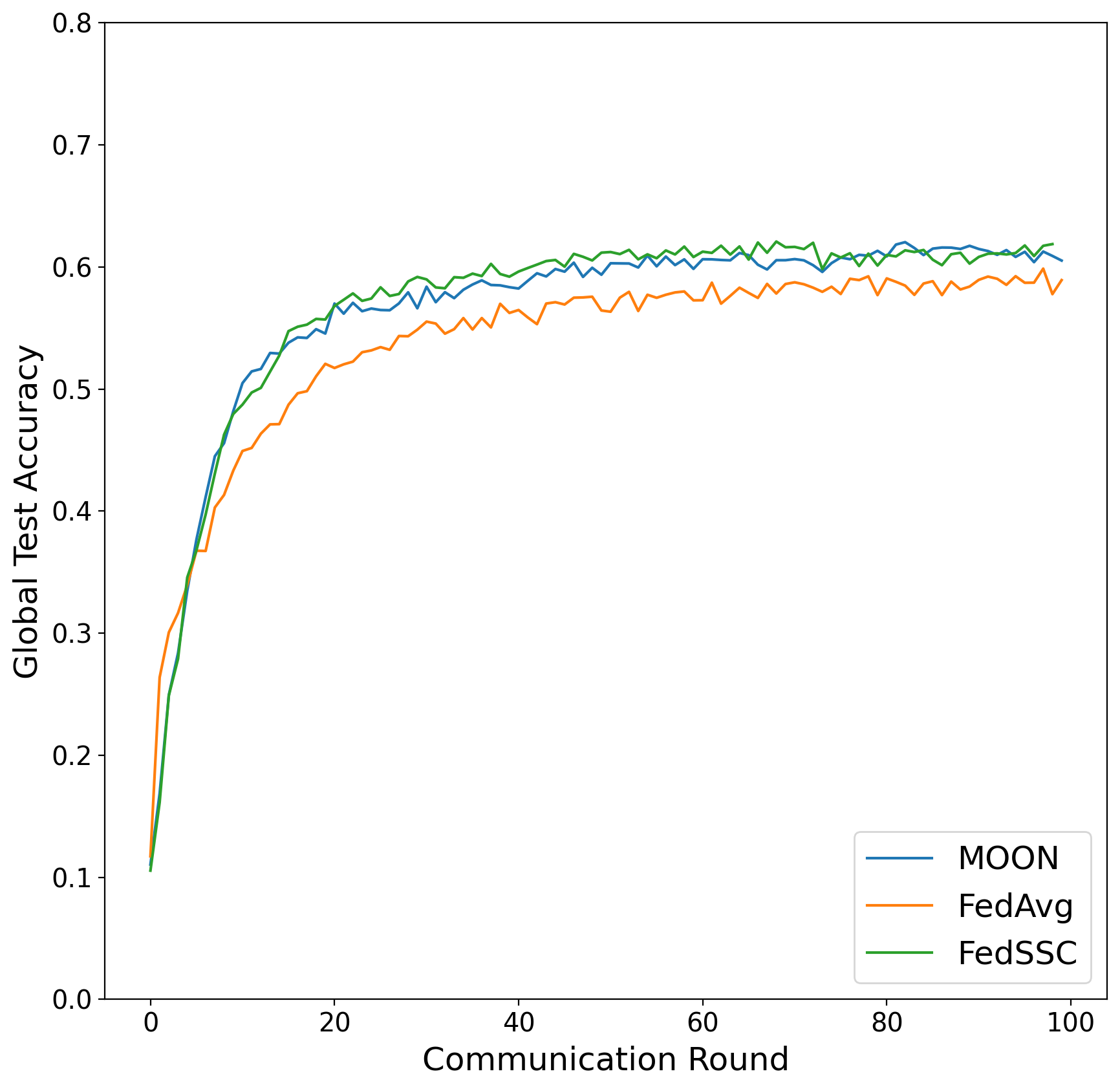}
  \caption{$\beta=0.2$}
  \label{fig:sub2A}
\end{subfigure}
\begin{subfigure}{0.24\linewidth}
  \centering
 \includegraphics[width=\linewidth]{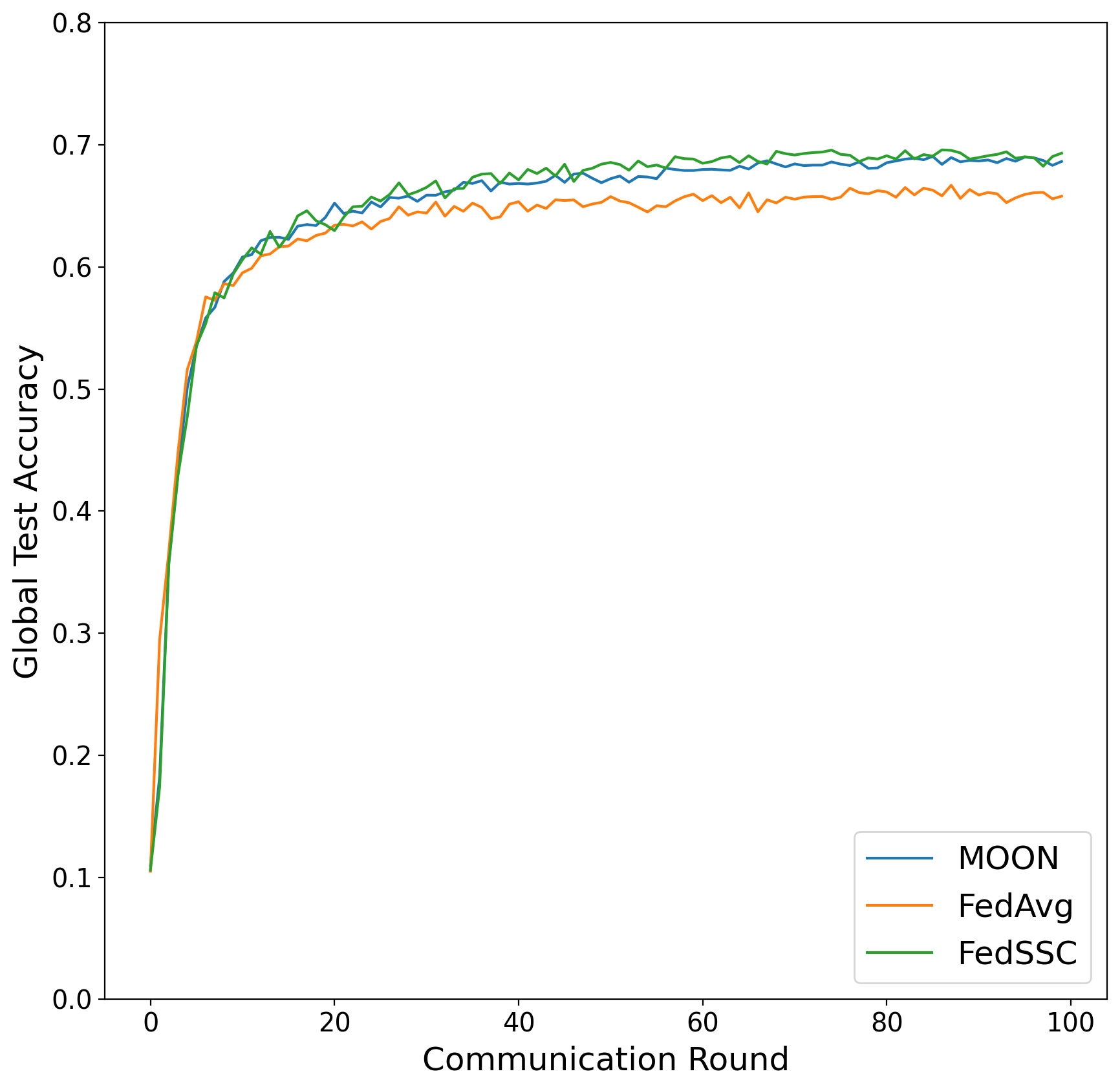}
  \caption{$\beta=0.5$}
\end{subfigure}
\begin{subfigure}{0.24\linewidth}
  \centering
  \includegraphics[width=\linewidth]{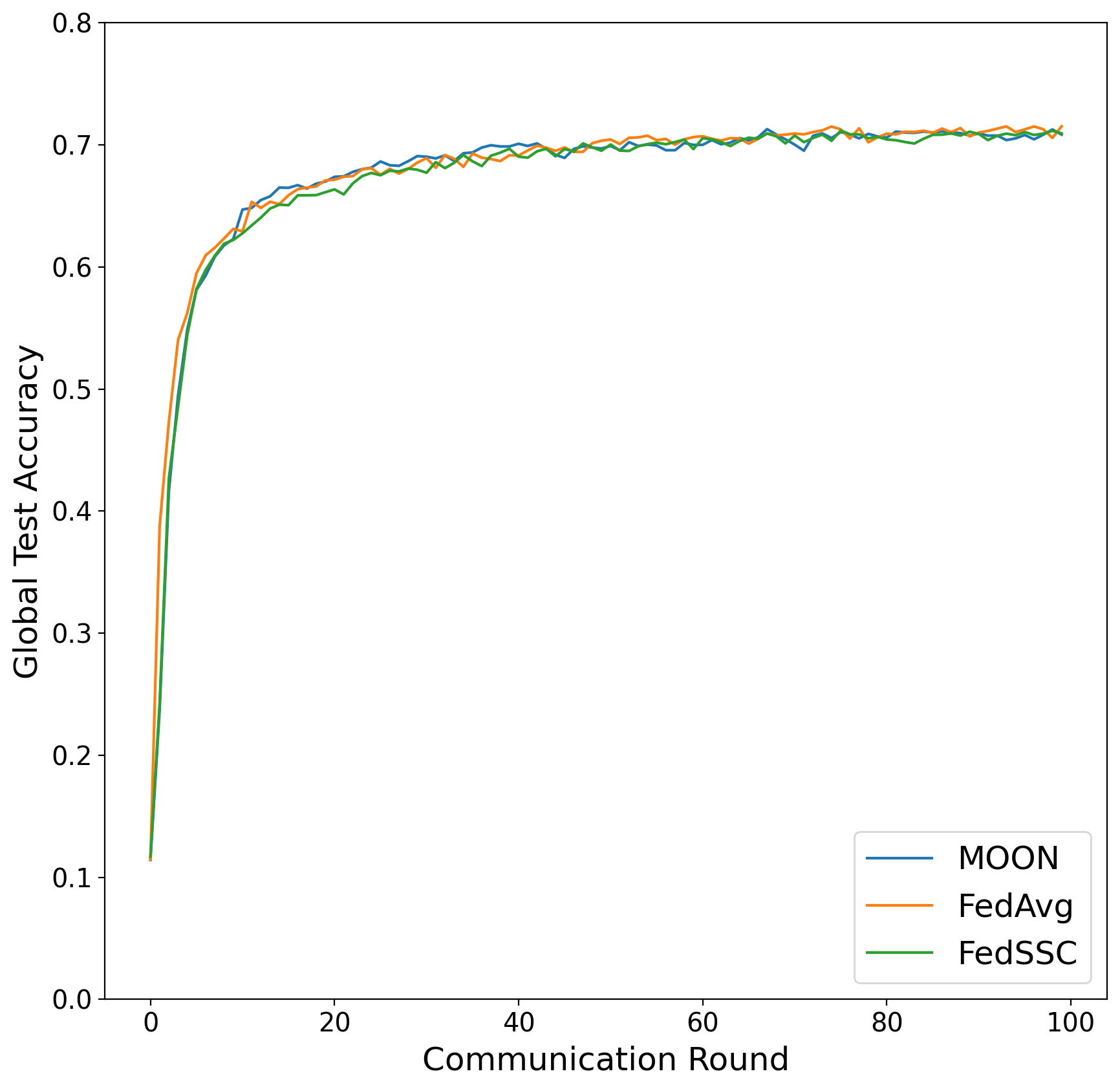}
  \caption{$\beta=1$}
  \label{fig:sub1A}
\end{subfigure}
\begin{subfigure}{0.24\linewidth}
  \centering
  \includegraphics[width=\linewidth]{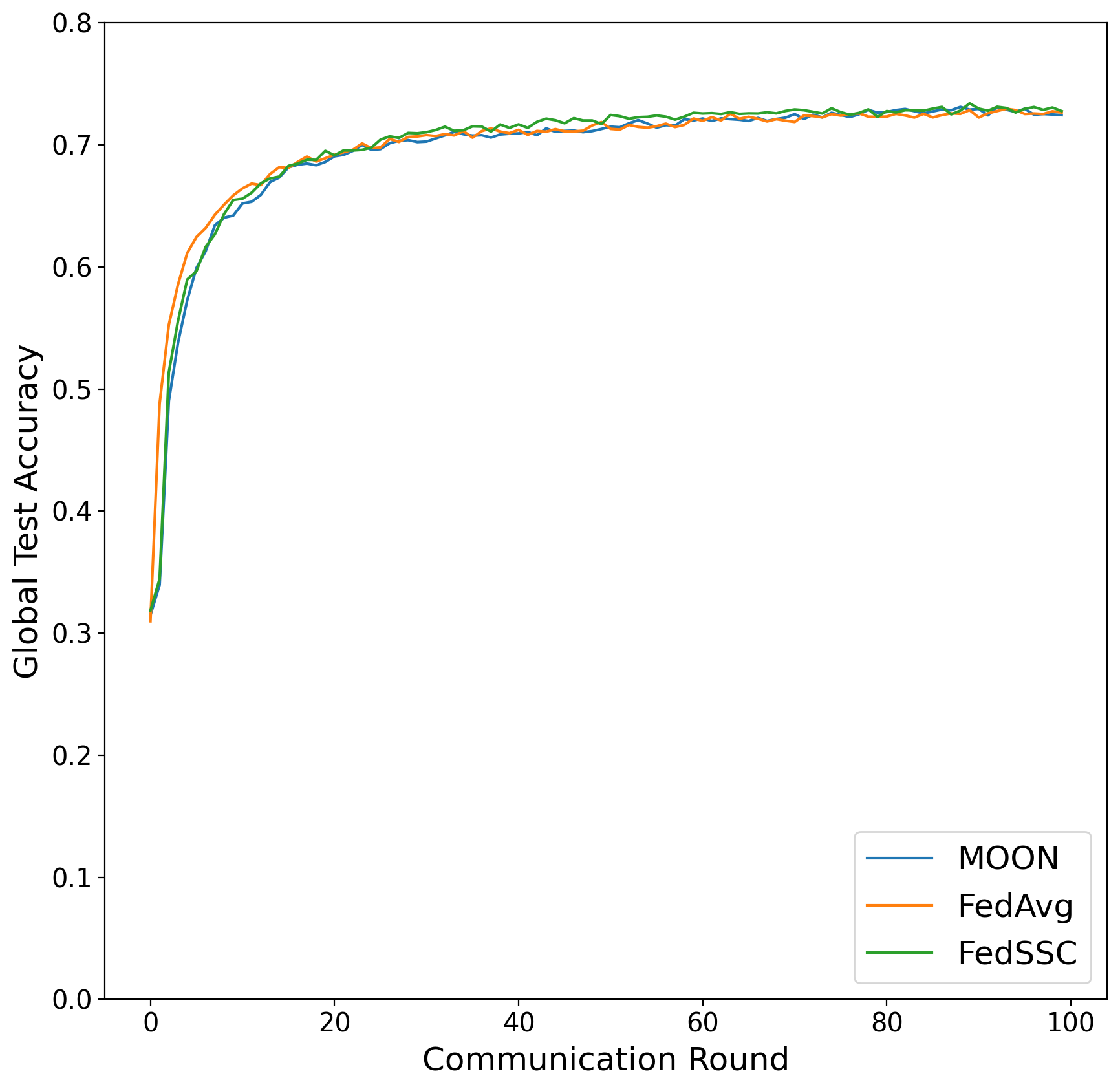}
  \caption{$\beta=5$}
\end{subfigure}
\caption{Performance comparison under different $\beta$ of non-IID scenarios.
}
\label{fig:beta_exp}
\end{figure*}

\subsection{Implementation Details}
\textbf{Main Differences Compared with MOON} Our approach is implemented as an extension to MOON with two main differences. In particular, we modify the loss function to include supervised contrastive loss for shared representations. Furthermore, we change the device-to-server communication to sharing class-wise average representations of the local model, in addition to the local model's weights. To avoid bias from limited data points, we only share the representation if the device has abundant samples for the corresponding class (i.e., more than 10 samples). Moreover, when distributing representations from the server to a device, we randomly sample $k$ representations for each class and take the average of them to represent the specific class. If the server has fewer than $k$ representations for a class, we will average everything we have to represent that class. 

\textbf{Model Architecture} Considering the time limit of this study, we used a simple CNN with two convolution layers, two max-pooling layers, and two fully connected layers as the encoder. After each convolution layer and fully connected layers, we have ReLU for the activation. 

\subsection{Experiment Setups}
To evaluate the performance of our method, we compared it with FedAvg and MOON, where MOON is expected to perform better than FedAvg in non-IID setting. In default, we set $\mu_{MOON}=5$ as it is the best parameter reported by the original paper \cite{moon}. Besides, we set the batch size to $64$, and use the SGD optimizer with a learning rate of $0.01$, a weight decay of $0.00001$, and a momentum of $0.9$. 

Furthermore, to mimic traditional two-stage contrastive learning, we decrease the weight of the global representation contrastive loss throughout the communication rounds. Specifically, we use the following formula to control its weight, where $\mu_{glob, s}$ is the initial weight, the $\mu_{glob, e}$ is the end weight, and the $T_{0}$ is the number of warmup rounds. In our experiment, we set $\mu_{glob, s} = 1$, $\mu_{glob, e} = 0.0001$, $T=100$, and $T_0=5$.

$$
\mu_{glob, i} = \mu_{glob, s} - \frac{1}{T-T_{0}} (\mu_{glob, s} - \mu_{glob, e})
$$

To simplify our experiments, we assume that no device will reject the server's request, and we will not encounter communication failures. In other words, all devices will participate in each communication round. In our default setting, we set the number of devices to 10. We utilized $1$ GPU and $10$ CPUs for each experimental setting. The training took less than 1 day to finish for all the experiments. 

\section{Results}
\begin{table}[h]
\label{results-summary}
\caption{Overall performance and efficiency for different methods.}
\begin{small}
\begin{sc}
\begin{tabular}{lcc}
\hline
Method & Top-1 Accuracy & Num of Comms (0.68 Acc)\\
\hline
FedAvg    & 0.658 & more than 100\\
MOON  & 0.686  & 61 \\
FedSSC    & \textbf{0.693} & \textbf{41}\\
\hline
\end{tabular}
\end{sc}
\end{small}
\end{table}

\subsection{Overall Performance}
In the default setting, our method \textsc{FedSSC} outperforms \textsc{FedAvg} by $3.4\%$ and \textsc{MOON} by $0.7\%$. The improvement compared to \textsc{FedAvg} is significant, while the increase from \textsc{MOON} is smaller. However, \textsc{FedSSC} reaches the $68\%$ level of accuracy at just $41$ communication rounds, while \textsc{MOON} needs $61$ rounds, and \textsc{FedAvg} cannot reach it within $100$ rounds. In summary, our approach performs better and is more efficient than previous methods.

\begin{figure*}[ht]
\centering
\begin{subfigure}{0.3\linewidth}
 \centering
 \includegraphics[width=0.9\linewidth]{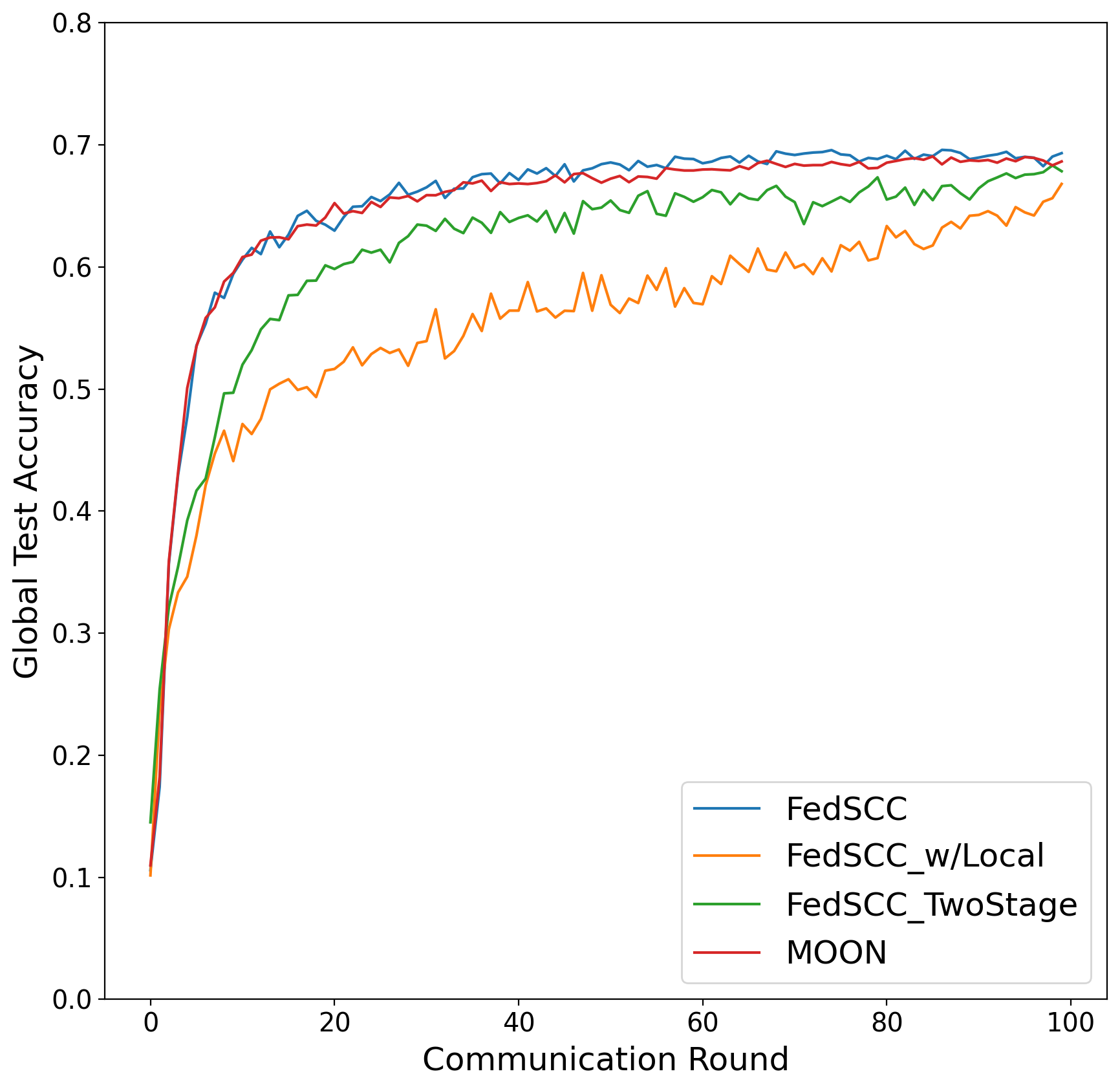}
\caption{Losses w/o model contrastive learning}
\label{fig:loss_exp_sub1}
\end{subfigure}
\begin{subfigure}{0.3\linewidth}
  \centering
 \includegraphics[width=0.9\linewidth]{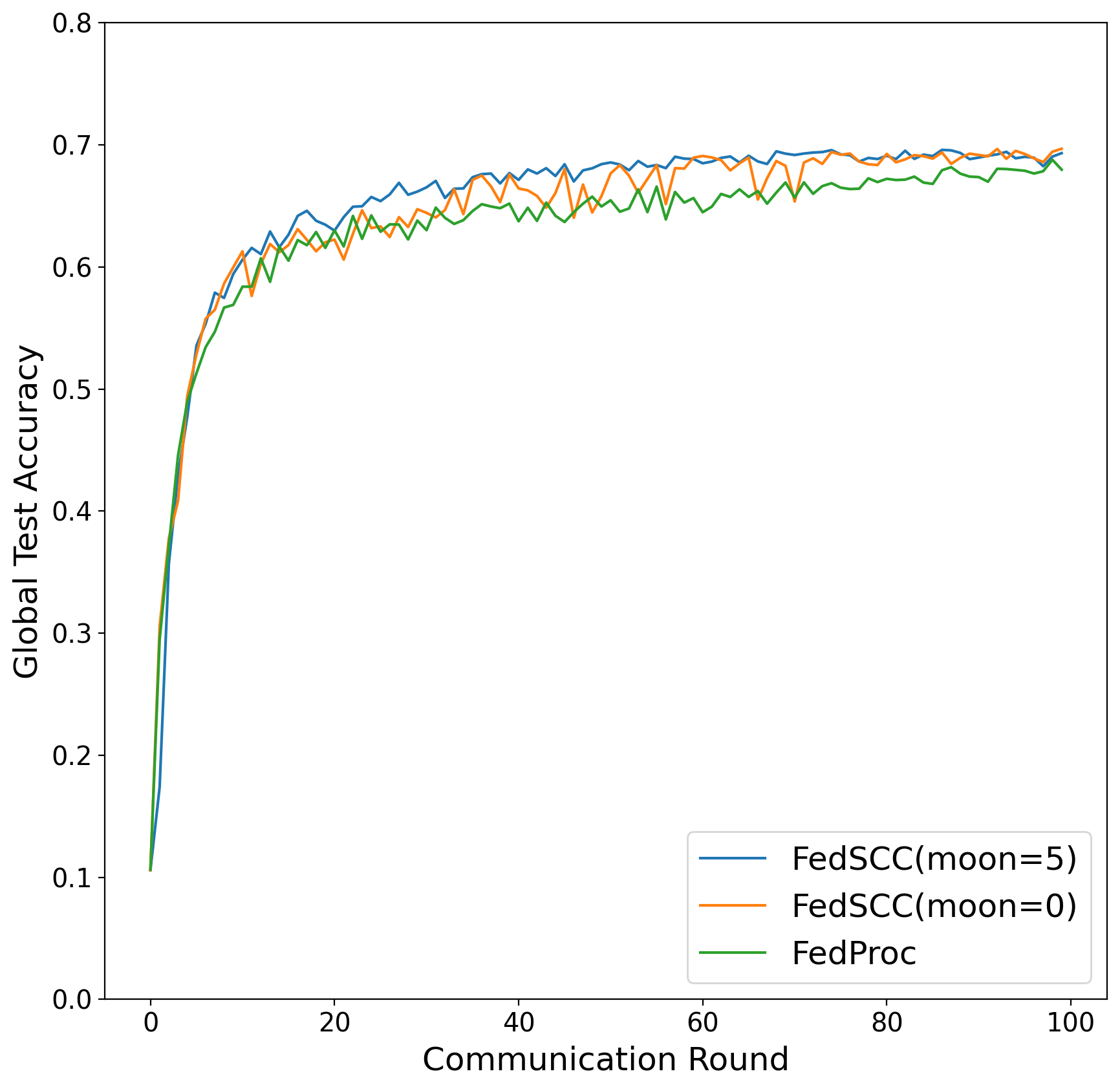}
\caption{Losses w/ local contrastive learning}
\label{fig:loss_exp_sub2}
\end{subfigure}
\begin{subfigure}{0.3\linewidth}
  \centering
 \includegraphics[width=0.965\linewidth]{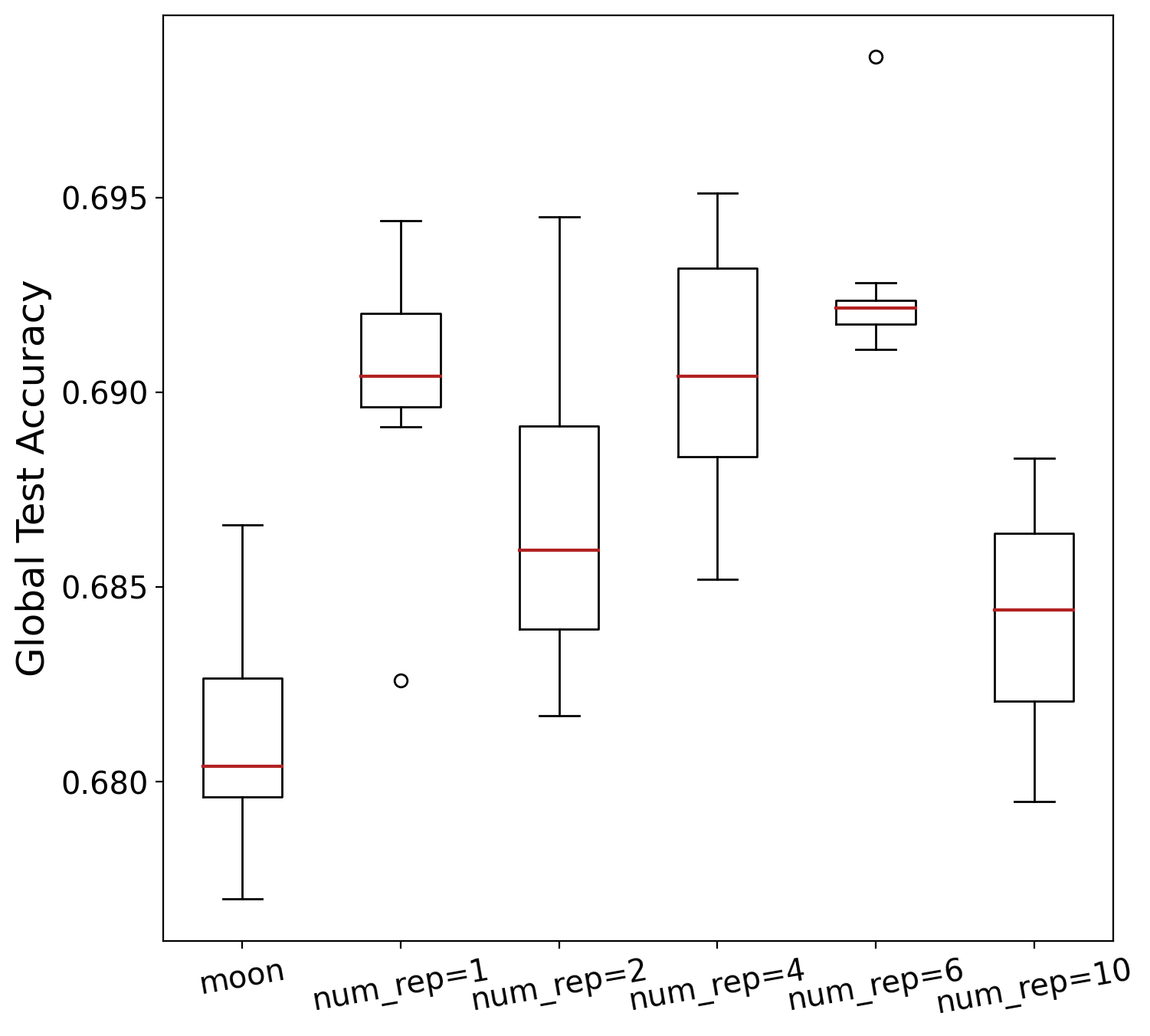}
  \caption{Different $k$ of presentations shared}
  \label{fig:loss_exp_sub3}
\end{subfigure}
\caption{Performance comparison for alternative loss components, where the blue line is our proposed approach. }
\label{fig:loss_exp}
\end{figure*}

\subsection{Different Non-IID Scenarios}
To better evaluate \textsc{FedSSC}, we compared its performance with other approaches under various non-IID scenarios, including $\beta \in \{0.2, 0.5, 1, 5\}$. From Figure \ref{fig:beta_exp}, we can see that as we increase $\beta$, the differences between the methods decrease since the heterogeneity is less severe. For both $\beta=0.2$ and $0.5$, \textsc{FedSSC} slightly outperforms \textsc{MOON} and is significantly better than \textsc{FedAvg}. Also, it converges faster than the others.   

\subsection{Alternative Loss}

Beyond our proposed loss, we did extensive experiments with other variants that could potentially help us learn a better representation. In particular, we tried to use two-stage contrastive learning, where we first train the encoder for certain rounds while freezing the classifier in the first 90 rounds and then train the classifier while freezing the encoder in the last 10 rounds. Furthermore, we experimented with another alternative by adding negative pairs and positive pairs from the same local batch. Figure \ref{fig:loss_exp_sub1} shows that our proposed approach outperforms the alternatives a lot, especially in the earlier rounds. This is understandable because learning encoder with only supervised contrastive learning would take longer rounds to converge than using supervised classification loss. However, it could potentially outperform our current approach if we train it for more rounds.

Furthermore, we attempted to remove the $l_{MOON}$ completely by setting $\mu_{MOON}=0$ and it achieves the same level of accuracy. From Figure \ref{fig:loss_exp_sub2}, we can see that $l_{MOON}$ takes an important role. If we remove it, the performance will drop even if we share representations from all devices (i.e., \textsc{FedProc} \cite{contrafl-1}). Moreover, we experimented with different numbers of shared representations $k$ with the default $\mu_{MOON}=0$. Figure \ref{fig:loss_exp_sub3} shows that sharing representations does boost the performance, although increasing $k$ doesn't make a big difference. Therefore, the experimental results demonstrate that $l_{MOON}$ and $l_{glob}$ are complementary to each other.

\section{Related Work}
Recently many methods have been proposed to improve model accuracy and data usage at a heterogeneous distribution environment, as it has long been a key problem for Federated Learning in many fields such as finance, medicine, and social media where participants don't want to share private data.

\textbf{Federated Learning}
Based on the work of FedAvg \cite{fedavg}, many methods have been proposed to alleviate the heterogeneous distribution problem. FedProx \cite{fedprox} adds an extra regularization term to push together local model weights and global model weights. SCAFFOLD \cite{scaffold} method uses variance reduction to correct the heterogeneity during local training. SphereFed \cite{sphere} makes use of a freezed classification head to increase the similarity between global and local feature space. Generally, most of the previous methods focus on bring together local and global models.


\textbf{Contrastive learning.} Methods such as SimCLR \cite{simclr} and MOCO \cite{moco} have become promising self-supervised approaches in Computer Vision in recent years. BYOL \cite{byol} and Simsiam\cite{simsiamese} have extended the idea of contrastive learning to have zero negative samples, while SupCon \cite{supcon} proposed a supervised contrastive learning approach.
some researchers have combined contrastive learning with federated learning to mitigate the heterogeneous distribution problem\cite{moon}, \cite{contrafl-2}, \cite{contrafl-1}. Our approach is improved upon the idea of MOON \cite{moon}, where positive pair is the local representation and global representation of the same image, and the negative pair is set to be the current local representation and the local representation at the previous communication round.

\textbf{Representation sharing}. Recently, representation sharing becomes another direction to solve the problem of heterogeneous data distribution. There are quite some research works dedicated in this direction, by sharing clients' image-level or class-level representation with others. FedProc \cite{contrafl-1} proposed sharing both the local features and local model weights. The local features are averaged in a class-wise manner, which has achieved better performance than MOON or FedProx on CIFAR-10 or CIFAR-100. Another recent work, FedPCL \cite{contrafl-2} applies individual-level feature sharing on MINIST dataset.

\section{Conclusion}
In summary, our work proposes to utilize contrastive learning and representation sharing to mitigate the non-IID problem. The experiments show that our method is orthogonal to other federated learning methods, and can outperform state-of-the-art models in typical settings. Both the accuracy and convergence speed can be apparently raised. Admittedly, more experiments are needed to test the availability and performance of our method with different settings of neural network structures and datasets.

\section{Contribution Statement}
All four authors contributed equally to this work. All members participated fully in reviewing literatures, coming up with model ideas, coding different models, tuning hyperparameters and writing up the final report.

\bibliographystyle{IEEEtran}
\bibliography{custom}

\begin{thebibliography}{10}
\providecommand{\url}[1]{#1}
\csname url@samestyle\endcsname
\providecommand{\newblock}{\relax}
\providecommand{\bibinfo}[2]{#2}
\providecommand{\BIBentrySTDinterwordspacing}{\spaceskip=0pt\relax}
\providecommand{\BIBentryALTinterwordstretchfactor}{4}
\providecommand{\BIBentryALTinterwordspacing}{\spaceskip=\fontdimen2\font plus
\BIBentryALTinterwordstretchfactor\fontdimen3\font minus
  \fontdimen4\font\relax}
\providecommand{\BIBforeignlanguage}[2]{{%
\expandafter\ifx\csname l@#1\endcsname\relax
\typeout{** WARNING: IEEEtran.bst: No hyphenation pattern has been}%
\typeout{** loaded for the language `#1'. Using the pattern for}%
\typeout{** the default language instead.}%
\else
\language=\csname l@#1\endcsname
\fi
#2}}
\providecommand{\BIBdecl}{\relax}
\BIBdecl

\bibitem{zhang2022fine}
L.~Zhang, L.~Shen, L.~Ding, D.~Tao, and L.-Y. Duan, ``Fine-tuning global model
  via data-free knowledge distillation for non-iid federated learning,'' in
  \emph{Proceedings of the IEEE/CVF Conference on Computer Vision and Pattern
  Recognition}, 2022, pp. 10\,174--10\,183.

\bibitem{zhu2021data}
Z.~Zhu, J.~Hong, and J.~Zhou, ``Data-free knowledge distillation for
  heterogeneous federated learning,'' in \emph{International Conference on
  Machine Learning}.\hskip 1em plus 0.5em minus 0.4em\relax PMLR, 2021, pp.
  12\,878--12\,889.

\bibitem{li2020federated}
T.~Li, A.~K. Sahu, A.~Talwalkar, and V.~Smith, ``Federated learning:
  Challenges, methods, and future directions,'' \emph{IEEE Signal Processing
  Magazine}, vol.~37, no.~3, pp. 50--60, 2020.

\bibitem{chen2022anomalous}
Y.~Chen, Z.~Guan, J.~Liu, W.~Yang, and H.~Wang, ``Anomalous layer-dependent
  lubrication on graphene-covered-substrate: Competition between adhesion and
  plasticity,'' \emph{Applied Surface Science}, p. 153762, 2022.

\bibitem{huang2022learn}
W.~Huang, M.~Ye, and B.~Du, ``Learn from others and be yourself in
  heterogeneous federated learning,'' in \emph{Proceedings of the IEEE/CVF
  Conference on Computer Vision and Pattern Recognition}, 2022, pp.
  10\,143--10\,153.

\bibitem{simclr}
\BIBentryALTinterwordspacing
T.~Chen, S.~Kornblith, M.~Norouzi, and G.~E. Hinton, ``A simple framework for
  contrastive learning of visual representations,'' \emph{CoRR}, vol.
  abs/2002.05709, 2020. [Online]. Available:
  \url{https://arxiv.org/abs/2002.05709}
\BIBentrySTDinterwordspacing

\bibitem{simsiamese}
\BIBentryALTinterwordspacing
X.~Chen and K.~He, ``Exploring simple siamese representation learning,''
  \emph{CVPR}, vol. abs/2011.10566, 2021. [Online]. Available:
  \url{https://arxiv.org/pdf/2011.10566.pdf}
\BIBentrySTDinterwordspacing

\bibitem{moon}
\BIBentryALTinterwordspacing
Q.~Li, B.~He, and D.~Song, ``Model-contrastive federated learning,''
  \emph{CoRR}, vol. abs/2103.16257, 2021. [Online]. Available:
  \url{https://arxiv.org/abs/2103.16257}
\BIBentrySTDinterwordspacing

\bibitem{fedavg}
\BIBentryALTinterwordspacing
H.~B. McMahan, E.~Moore, D.~Ramage, and B.~A. y~Arcas, ``Federated learning of
  deep networks using model averaging,'' \emph{CoRR}, vol. abs/1602.05629,
  2016. [Online]. Available: \url{http://arxiv.org/abs/1602.05629}
\BIBentrySTDinterwordspacing

\bibitem{fedprox}
\BIBentryALTinterwordspacing
A.~K. Sahu, T.~Li, M.~Sanjabi, M.~Zaheer, A.~Talwalkar, and V.~Smith, ``On the
  convergence of federated optimization in heterogeneous networks,''
  \emph{CoRR}, vol. abs/1812.06127, 2018. [Online]. Available:
  \url{http://arxiv.org/abs/1812.06127}
\BIBentrySTDinterwordspacing

\bibitem{contrafl-1}
\BIBentryALTinterwordspacing
X.~Mu, Y.~Shen, K.~Cheng, X.~Geng, J.~Fu, T.~Zhang, and Z.~Zhang, ``Fedproc:
  Prototypical contrastive federated learning on non-iid data,'' vol.
  abs/2109.12273, 2021. [Online]. Available:
  \url{https://arxiv.org/abs/2109.12273}
\BIBentrySTDinterwordspacing

\bibitem{scaffold}
\BIBentryALTinterwordspacing
S.~P. Karimireddy, S.~Kale, M.~Mohri, S.~J. Reddi, S.~U. Stich, and A.~T.
  Suresh, ``Scaffold: Stochastic controlled averaging for federated learning,''
  \emph{PMLR}, vol. abs/1910.06378, 2019. [Online]. Available:
  \url{https://arxiv.org/abs/21910.06378}
\BIBentrySTDinterwordspacing

\bibitem{sphere}
\BIBentryALTinterwordspacing
X.~Dong, S.~Q. Zhang, A.~Li, and H.~Kung, ``Spherefed: Hyperspherical federated
  learning,'' vol. abs/2207.09413, 2022. [Online]. Available:
  \url{https://arxiv.org/abs/2207.09413}
\BIBentrySTDinterwordspacing

\bibitem{moco}
\BIBentryALTinterwordspacing
K.~He, H.~Fan, Y.~Wu, S.~Xie, and R.~Girshick, ``Momentum contrast for
  unsupervised visual representation learning,'' \emph{CVPR}, vol.
  abs/1911.05722, 2020. [Online]. Available:
  \url{https://arxiv.org/abs/1911.05722}
\BIBentrySTDinterwordspacing

\bibitem{byol}
\BIBentryALTinterwordspacing
J.-B. Grill, F.~Strub, F.~Altché, C.~Tallec, P.~H. Richemond, E.~Buchatskaya,
  C.~Doersch, B.~A. Pires, Z.~D. Guo, M.~G. Azar, B.~Piot, K.~Kavukcuoglu,
  R.~Munos, and M.~Valko, ``Bootstrap your own latent a new approach to
  self-supervised learning,'' vol. abs/2006.07733, 2020. [Online]. Available:
  \url{https://arxiv.org/abs/2006.07733}
\BIBentrySTDinterwordspacing

\bibitem{supcon}
\BIBentryALTinterwordspacing
P.~Khosla, P.~Teterwak, C.~Wang, A.~Sarna, Y.~Tian, P.~Isola, A.~Maschinot,
  C.~Liu, and D.~Krishnan, ``Supervised contrastive learning,'' \emph{NeurIPS
  2020}, vol. abs/2004.11362, 2020. [Online]. Available:
  \url{https://arxiv.org/abs/2004.11362}
\BIBentrySTDinterwordspacing

\bibitem{contrafl-2}
\BIBentryALTinterwordspacing
Y.~Tan, G.~Long, J.~Ma, L.~Liu, T.~Zhou, and J.~Jiang, ``Federated learning
  from pre-trained models: A contrastive learning approach,'' vol.
  abs/2209.10083, 2022. [Online]. Available:
  \url{https://arxiv.org/abs/2209.10083}
\BIBentrySTDinterwordspacing

\end{thebibliography}

\vspace{12pt}

\end{document}